\documentclass[10pt,twocolumn,letterpaper]{article}

\usepackage{cvpr}
\usepackage{times}
\usepackage{epsfig}
\usepackage{graphicx}
\usepackage{amsmath}
\usepackage{amssymb}
\usepackage{sidecap}
\usepackage{balance}

\DeclareMathOperator*{\argmax}{argmax} 

\usepackage[pagebackref=true,breaklinks=true,letterpaper=true,colorlinks,bookmarks=false]{hyperref}

\cvprfinalcopy 

\def\cvprPaperID{3684}

\ifcvprfinal\fi
\begin{document}

\title{Towards Robust Curve Text Detection with Conditional Spatial Expansion}

\author{Zichuan Liu$^1$, Guosheng Lin$^1$, Sheng Yang$^1$, Fayao Liu$^2$, Weisi Lin$^1$ and Wang Ling Goh$^1$\\
	$^{1}$Nanyang Technological University, Singapore\\
	$^{2}$University of Adelaide, Australia\\
	{\tt\small \{zliu016, syang014\}@e.ntu.edu.sg}, {\tt\small \{gslin, wslin, ewlgoh\}@ntu.edu.sg},
	{\tt\small fayaoliu@gmail.com}
}

\maketitle
\thispagestyle{empty}

\begin{abstract}
    It is challenging to detect curve texts due to their irregular shapes and varying sizes. In this paper, we first investigate the deficiency of the existing curve detection methods and then propose a novel Conditional Spatial Expansion (CSE) mechanism to improve the performance of curve text detection. Instead of regarding the curve text detection as a polygon regression or a segmentation problem, we treat it as a region expansion process. Our CSE starts with a seed arbitrarily initialized within a text region and progressively merges neighborhood regions based on the extracted local features by a CNN and contextual information of merged regions. The CSE is highly parameterized and can be seamlessly integrated into existing object detection frameworks. Enhanced by the data-dependent CSE mechanism, our curve text detection system provides robust instance-level text region extraction with minimal post-processing. The analysis experiment shows that our CSE can handle texts with various shapes, sizes, and orientations, and can effectively suppress the false-positives coming from text-like textures or unexpected texts included in the same RoI. Compared with the existing curve text detection algorithms, our method is more robust and enjoys a simpler processing flow. It also creates a new state-of-art performance on curve text benchmarks with F-score of up to 78.4$\%$.
\end{abstract}
\section{Introduction}

\begin{figure}[]
	\centering
	\includegraphics[width=\linewidth]{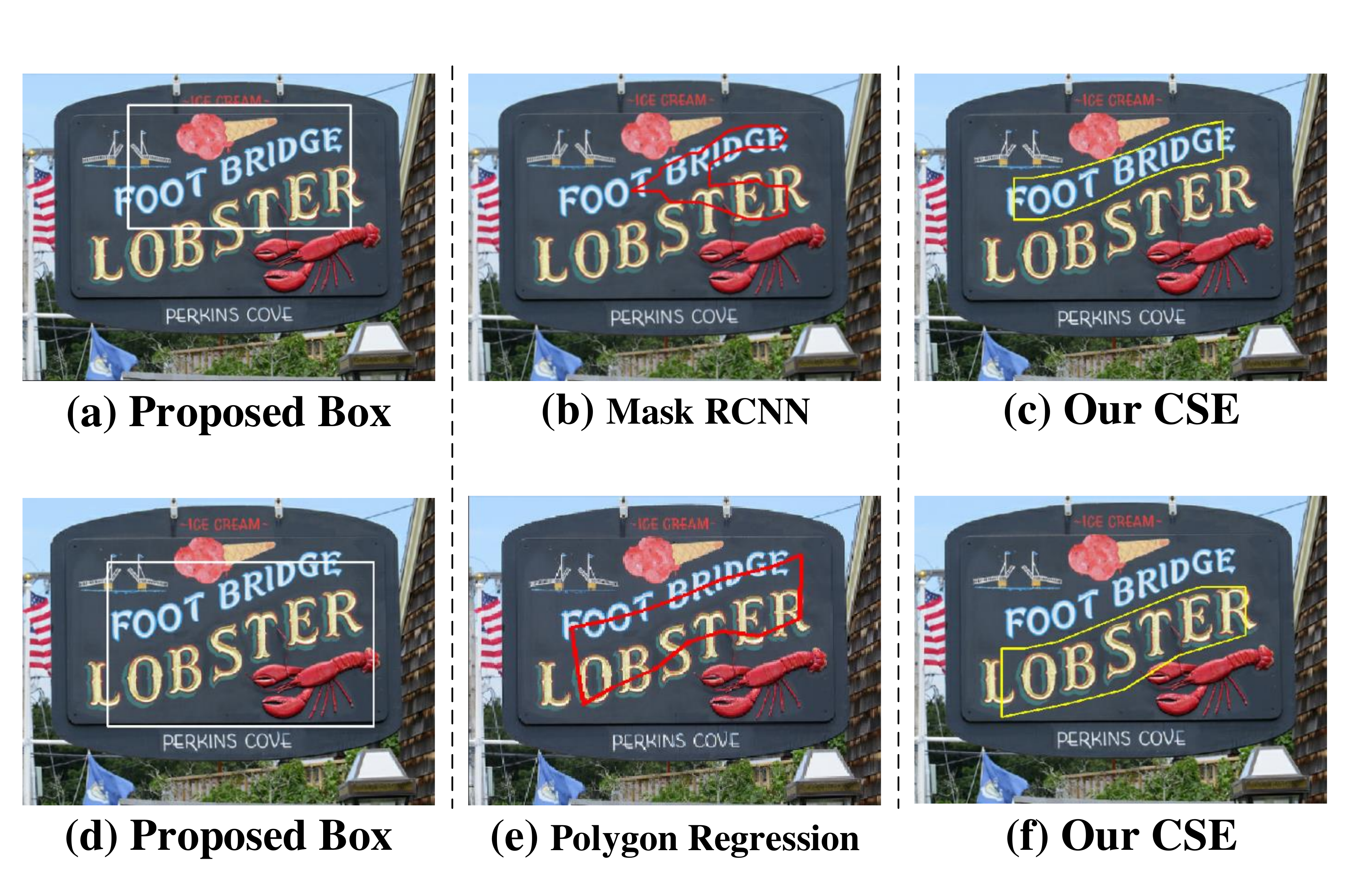}
	\label{fig:intro_demo}
	\caption{Problems of existing curve text detection methods: Two region proposals shown in (a) and (d) inevitably include unexpected texts since they are closed to each other. Thus, it causes failures for Mask RCNN based methods and polygon regression methods demonstrated in (b) and (e). Our CSE method demonstrated in (c) and (f) shows strong robustness to this situation and brings significant performance gain.}
\end{figure}

In recent years, great progress has been made in text detection. The performance has been enhanced by the advanced object detection and segmentation frameworks based on Neural Networks. Although detecting words or text lines with different sizes and orientations has been well tackled by recently proposed methods \cite{lyu2018multi, liu2018mcn,tian2016detecting,shi2017detecting,liu2018cpn}, detecting curve texts remains a challenging problem. \par

The main challenges of curve text detection come from irregular shapes and highly varying orientations. The traditional bounding box representation does not scale well in the curve scenario since one box may cover multiple text objects. Therefore, the recently proposed curve text detection algorithms \cite{yuliang2017detecting,lyu2018eccv,zhan2019} follow a two-stage detect-and-refine approach to generate elaborated polygons or boundaries. In these methods, a CNN based text detector is applied to locate the regions containing texts, and then a segmentation or polygon regression algorithm is performed on these regions to produce a tight polygon or boundary. Both methods highly depend on the accurate region proposal provided by the text detector. They prefer a proposed region with only one targeted object included which reduces the ambiguity of the sampled features. Although an oriented box regression is preferable, it often fails in the curve text scenario \cite{ma2018arbitrary,jiang2017r2cnn}. The recently proposed curve text detection methods turn to predict horizontal rectangles in the first stage, which is inevitably affected non-targeted texts in the sampled regions. 

Specifically, both segmentation and regression based methods can be disrupted by the unexpected texts included in the same box region. The segmentation based methods can fail to distinguish the targeted text from the others and misclassify the unexpected texts or text-like patterns as positive, shown in Fig. \ref{fig:intro_demo} (b). The regression based methods can produce incorrect boundaries by indistinguishably considering all texts as one object. Moreover, as shown in Fig. \ref{fig:intro_demo} (e), the regression results (produced by predicting an offset of a proposed box region) are highly coupled with previous stage box proposals. A poor box proposal greatly affects the final polygon which degrades the performance. \par

To tackle the problem mentioned above, we propose a novel Conditional Spatial Expansion (CSE) mechanism, which acts as a second-stage component applied in the widely adopted two-stage detection workflow. Our method is derived from conditional modeling of dependency between an interior point (called a seed) and the rest parts of a text instance. The CSE can be seen as a conditional prediction process which retrieves an instance level text region by seeding and expanding. Starting with an arbitrary interior point (seed) of a text region, CSE selectively expands its territory based on location observation of the image patches and the context inferred from merged regions. Compared with the segmentation based method, our CSE is extremely discriminative especially when texts are close to each other as demonstrated in Fig. \ref{fig:intro_demo} (c). It provides a controllable approach to extract an expected text region with minimum efforts of post-processing. On the other hand, our CSE is highly flexible since a seed can be specified at any location within the targeted text region. Compared with the polygon regression methods, the seeding-and-expanding paradigm has less coupling with the previous text detector. Base on a coarse region proposal, our CSE is directly applied on the spatial features produced by the backbone, which preserves all the spatial information and will not be affected by the imperfect region proposals, shown in Fig. \ref{fig:intro_demo} (f). The experiments show that our method outperforms the existing curve text detection methods on public benchmarks. The contributions of this work are summarized as follows:
\begin{itemize}
	\item The curve text detection is formulated as a conditional region expansion problem, which initializes a seed within a text region and then progressively retrieves targeted object by region expansion;
	\item The spatial dependency between the seed and the rest part of an object is modeled by a parameterized Conditional Spatial Expansion mechanism, which allows us to selectively extract a text region indicated by a seed with high area precision;
	\item Our CSE acts as a second-stage text extractor which can be seamlessly integrated into existing object detection workflows; 
	\item The arbitrariness of seed's location and high spatial selectivity of our method reduce coupling with the previous detector and thus provide flexible and robust boundary prediction;
	\item Our method outperforms the existing curve text detection methods on public curve text datasets with F-measurement of 80.2$\%$ on Total-Text \cite{ch2017total} and 78.4$\%$ on CTW-1500 \cite{yuliang2017detecting}.  
\end{itemize}

\section{Related Works}

\subsection{Quadrilateral Text Detection}
In the quadrilateral text detection, the ground-truths are constrained to a rectangle or a quadrilateral. Base on the types of targets to be retrieved, text detection methods can be categorized as detection based methods and segmentation based methods. \par

The detection beased method follows the object detection frameworks \cite{girshick2015fast,ren2015faster,liu2016ssd,redmon2016you,liu2018fots,he2018end} driven by Convolutional Neural Networks (CNNs) \cite{lecun1998gradient}. TextBoxes \cite{liao2017textboxes} adopts SSD as a base detector and handles variation of aspect ratios of text instance by elaborated reference box design. As the variants of the faster RCNN \cite{redmon2016you}, the Rotation Region Proposal Network (RRPN) \cite{ma2018arbitrary} and Rotational Region CNN (R$^2$CNN) \cite{jiang2017r2cnn} are designed to detect arbitrarily oriented texts in a two-stage manner. In addition, EAST \cite{zhou2017east} and DeepReg \cite{he2017deep} are proposed to directly regress the geometry of a text instance.\par

The segmentation based methods are mostly designed to extract long text lines in an image. They interpret text detection as a semantic segmentation problem which has been well addressed by the Fully Convolutional Neural Networks (FCNs) \cite{long2015fully,ding2018cvpr,ding2019semantic,zhan2018}. Zhang \emph{et al.} \cite{zhang2016multi} combines FCN and MSER \cite{huang2014robust} to recognize text blocks and then extract corresponding characters. Yao \emph{et al.} \cite{yao2016scene} applies FCN to predict multiple properties of texts, such as text regions and orientations, to extract the target text regions. To distinguish adjacent text instances, the component segmentation method \cite{liu2018mcn,liu2018cpn,wu2017self,deng2018pixellink,xue2018} is proposed, where a text region is broken into several components, which will be combined into different instances by data-driven clustering, inter-node communication or post-processing. \par

\subsection{Curve Text Detection}

Although the methods reviewed above have succeeded in the quadrilateral text detection, most of them cannot scale well in the case of arbitrary text shape. New representations and detection framework are proposed for this task. Liu \emph{et al.} \cite{yuliang2017detecting} propose Transverse and Longitudinal Offset Connection (TLOC) method based on Faster RCNN and Recurrent Neural Networks (RNNs) to directly regress the polygon shape of text regions. Mask Text-Spotter \cite{lyu2018eccv} regards the curve text detection as an instance segmentation problem and applies the Mask-RCNN to produce boundaries of text instances. TextSnake \cite{long2018eccv} adopts FCN as the base detector and extract text instance by detecting and assembling local components. \par

Most of the existing curve text detection methods are potentially modeling the posterior probability between observed image patches and the ground-true foreground labels. Instead, our method captures the dependency between arbitrary image patches and the rest of the text regions. The proposed modeling is naturally robust to ambiguity caused by multiple text instances included in one RoI. Moreover, our CSE considers more local details and thus can produce more elaborated text boundaries. \par

\section{Method}

\subsection{Overview}

Our method retrieves an instance level text region by seeding and then expanding. A seed uniquely indicates an object and can be arbitrarily initialized at the interior of an object region. Seeding is to select a location within an object from which the corresponding object region is extracted by expanding. As will be illustrated in Sect. \ref{sect:seeding}, a seed is initialized by an object detector with a predicted box center. Starting with a seed, the expanding is conducted by selectively merging adjacent sub-regions to form a targeted object region. As shown in Fig. \ref{fig:modeling}, the sub-regions are abstracted as feature points or nodes, which are sampled from the input image at discrete locations. They are organized as a grid and locally assigned an expanding indicator $y \in \mathbb{R}^5$ to represent the merging direction to neighborhood nodes. Five entries of $y$ denote the probabilities of all possible merging directions ($to-bottom$, $to-right$, $to-left$, $to-top$ and $none$). A node will be merged into an existing object region indicated by a seed if its major merging direction is pointing to its neighborhoods that already belong to the object region. An instance level object boundary can be easily produced by mapping all the positive nodes back to the original image and extracting the contour. \par

\subsection{Modeling}

\begin{figure}
	\includegraphics[width=0.45\textwidth]{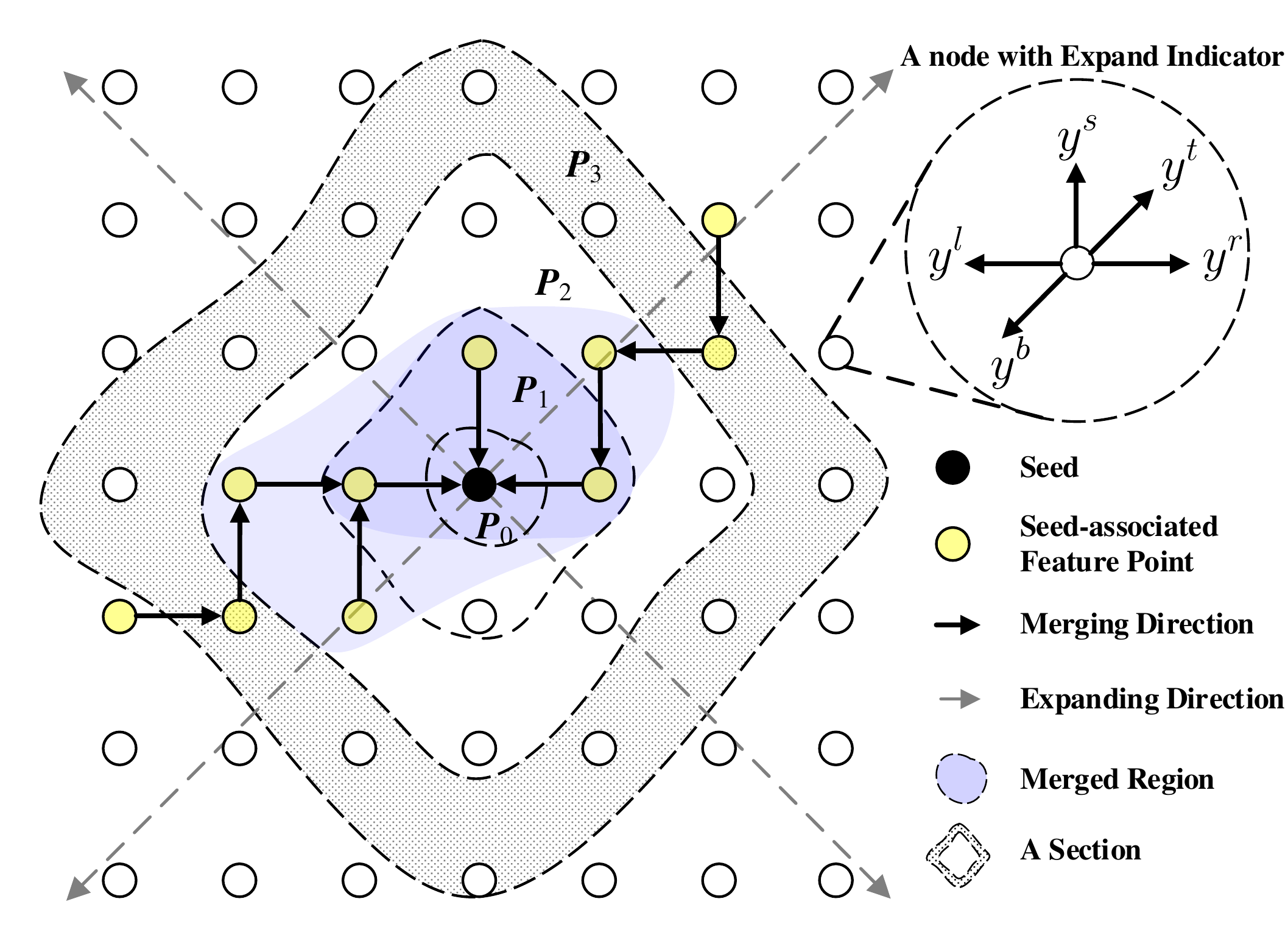}
	\centering
	\caption{Our CSE inside-out explores the expanding indicator of every node and merges nodes with merging direction pointing to nodes that are already in the object region.}
	\label{fig:modeling}
\end{figure}

The seeding-and-expanding paradigm provides a flexible and controllable way to extract object regions with minimal post-processing. It also reduces the performance coupling with the first-stage detector by allowing arbitrary initialization of a seed node. However, with different seed locations, the dynamics of the region expansion are different. For a specific node, the state of expanding indicator varies with the seed's location and also depends on the expanding indicators of its neighborhoods. Therefore, obtaining the expanding indicators should be regarded as a conditional prediction problem. Centered at a seed shown in Fig. \ref{fig:modeling}, we divide the region into sections (defined as nodes with the same minimum number of steps to a seed) using a set of contours. We assume that the expanding indicators of nodes $P_k$ within the $k$-th section are mutually independent and their states depend only on the current spatial feature $X_k$ and the states of nodes within previous sections $P_{k-1}, P_{k-2}, \cdots, P_0$. The optimal estimator of $Y_k$ can be represented by
\begin{align}
\label{eq:pr}
\hat{Y_k} = \argmax_{Y_k}{\Pr(Y_k|Y_{k-1}, \cdots, Y_{0};X_k, \cdots, X_0)},
\end{align} 
which maximizes the posterior probability of $Y_k$ when observing spatial features $X_{(\cdot)} := \{x(p)|p\in P_{(\cdot)}\}$ and indicators of previous nodes $Y_{(\cdot)} := \{y(p)|p\in P_{(\cdot)}\}$. This conditional modeling allows the region expansion to be adaptable with an arbitrarily initialized seed location. Also, it effectively differentiates expected object from the others by considering the context derived from a seed. Moreover, independence assumption among node in the same section results in a dendritic Conditional Spatial Expansion process with high-level parallelism. \par

\subsection{Conditional Spatial Expansion}

\begin{figure*}
	\includegraphics[width=0.9\textwidth]{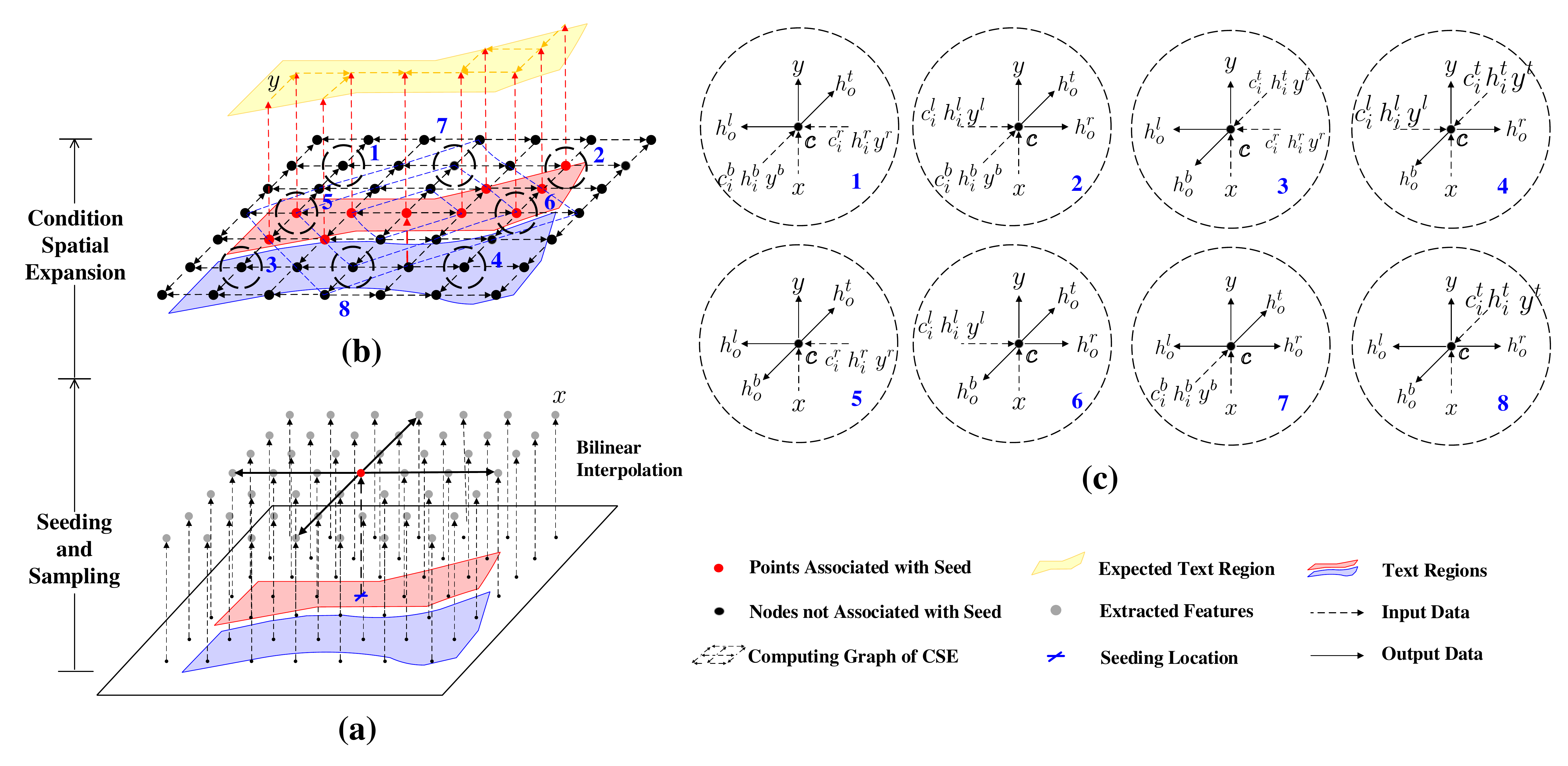}
	\centering
	\caption{Given a seed located at the interior of a text region, we expand a grid with $S\times S$ points and sample the feature produced by backbone at these locations using bilinear interpolation. The CSE computation starts with the seed and spreads to the adjacent feature nodes. Each node takes the outputs and the hidden states of previous nodes as input and produces new hidden state and output.}
	\label{fig:cse}
\end{figure*}

To estimate conditional probability illustrated in Eq. \ref{eq:pr}, we develop a highly parameterized Condition Spatial Expansion (CSE) mechanism. Given a seed inside an object region, we construct its neighborhood feature points by expanding a $S\times S$ grid and then sampling the features produced by the backbone network using bilinear interpolation, as shown in Fig. \ref{fig:cse} (a). Starting with the seed node, our CSE explores every single node inside-out and computes corresponding $y$ and the transition vectors $h_o = [h_o^b, h_o^r, h_o^l, h_o^t]^T$ to its neighborhoods based on the current sampled feature $x \in \mathbb{R}^{d_x}$, local state $c \in \mathbb{R}^d$ and transition vectors $h_i^b, h_i^r, h_i^l, h_i^t \in \mathbb{R}^{d}$ coming from the adjacent feature points. The transition vectors encode the position sensitive information which helps the CSE to be aware of the relative location of the current node to the seed. Depending on the relative position to the seed, the inputs and outputs for the nodes are illustrated in Fig. \ref{fig:cse} (c). For a node in $P_k$, our CSE only takes the $c$ and $h$ as inputs from $P_{k-1}$ and output new $h$ to $P_{k+1}$. This constructs an inference process originated from the seed which propagates the contextual information among the grid in a dendritic manner. The computation of the nodes in the same section is independent and thus can be fully parallelized on GPUs. The computation complexity is linear to the side of the grid, which is computationally efficient.\par

\begin{figure}[]
	\centering
	\includegraphics[width=0.9\linewidth]{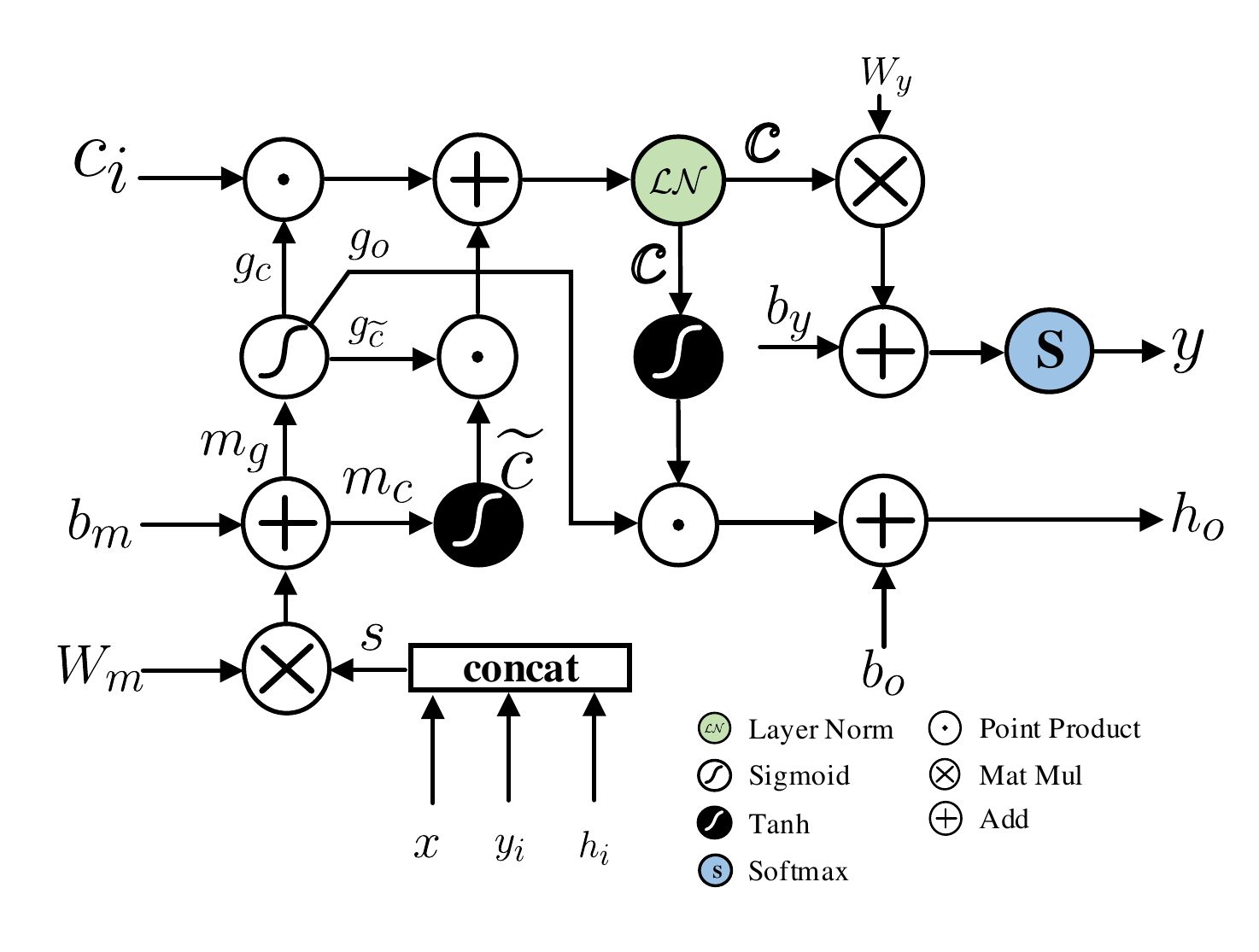}
	\caption{Computing Graph inside a Node.}
	\label{fig:cse_unit}
\end{figure}

Inside a specific node, the computation is illustrated by a computing graph shown in Fig \ref{fig:cse_unit}. Without loss of generality, we denote all possible inputs from neighborhood nodes by $c_i \in \mathbb{R}^{4d}$, $h_i \in \mathbb{R}^{5d}$, and $y_i \in \mathbb{R}^{20}$, which are represented by
\begin{align}
& c_i = [c_i^b, c_i^r, c_i^l, c_i^t]^T, \\
& h_i = [h_i^c, h_i^b, h_i^r, h_i^l, h_i^t]^T, \\
& y_i = [y_i^b, y_i^r, y_i^l, y_i^t]^T, 
\end{align}
where $c_i^{(\cdot)} \in \mathbb{R}^d$, $h_i^{(\cdot)} \in \mathbb{R}^d$ and $y_i^{(\cdot)} \in \mathbb{R}^5$ denotes the local states, transition vectors and the expanding indicators of neighborhood nodes \footnote{The superscripts indicate the relative position to the current node, b-bottom, r-right, l-left, t-top and c-current.}. Here, only parts of $c_i$, $h_i$ and $y_i$ are available according to the relative position to the seed node \footnote{The non-zero inputs are indicated in Fig. \ref{fig:cse} (c)}, and the rest will be set to zeros. Particularly, $h_i^c$ is defined to uniquely indicate the seed node. Other than $h_i^c$ of a seed which is learned by backpropagation, we explicitly set the $h_i^c$ of other node to zeros.\par

From the current observed feature $x$, transition vectors $h_i$ and the predicted expanding indicators of neighborhoods $y_i$, we compute a candidate local state $\widetilde{c}$ by
\begin{align}
& \widetilde{c} = \tanh (W_c \times s + b_c), \\
& s = [x, y_i, h_i]^T,
\end{align}
where $\times$ represents the matrix multiplication, $W_c \in \mathbb{R}^{d\times (d_x + 5d + 20)}$ and $b_c \in \mathbb{R}^d$ denote weights and bias of linear transform before a $\tanh$ activation. We apply the gating mechanism \cite{hochreiter1997long} to combine the local state $c_b$, $c_r$, $c_l$ and $c_t$ from the neighborhoods with the current candidate state $\widetilde{c}$ to obtain the local state of current node $c$, which is formulated as
\begin{align}
c = \delta (c_b \cdot g_c^b + c_r \cdot g_c^r + c_l \cdot g_c^l + c_t \cdot g_c^t + \widetilde{c} \cdot g_{\widetilde{c}}) \label{eq:c},
\end{align}
where $\delta$ denotes the layer normalization operator \cite{ba2016layer}, $\cdot$ is the element-wise multiplication, and $g_b$, $g_r$, $g_l$, $g_t$, $g_c$ represent the outputs of gating function which can be further illustrated by
\begin{align}
& g_c = [g_{c_i}^b, g_{c_i}^r, g_{c_i}^l, g_{c_i}^t]^T = \sigma (W_{g_{c_i}} \times s + b_{g_{c_i}}), \\
& g_{\widetilde{c}} = \sigma (W_{g_{\widetilde{c}}} \times s + b_{g_{\widetilde{c}}}).
\end{align}
Here, $W_{g_{c_i}}$, $b_{g_{c_i}}$ and $W_{g_{\widetilde{c}}}$, $ b_{g_{\widetilde{c}}}$ are defined as the weight matrix and bias to map $s$ into corresponding gating vectors $g_{c_i}$ and $g_{\widetilde{c}}$. Since the local state $c$ is essentially the weighted sum of state values of the previous nodes, the values of $c$ increases exponentially with $k$ in our two-dimensional scenario. This significantly harms the numerical stability in both training and testing phases. Thus, the layer normalization technique is essential for the CSE to ensure the convergence of training and prevent overflow in testing. Finally, the expanding indicator $y$ and the output transition vectors $h_o$ are derived from the local state $c$, which are illustrated as follows
\begin{align}
& g_o =  \sigma (W_{g_o} \times s + b_{g_o}), \\
& h_o = [h_o^b, h_o^r, h_o^l, h_o^t]^T = \tanh (c) \cdot g_o + b_o, \\
& y = softmax (W_y \times c + b_y),
\end{align}
where $W_{g_o}$ and $b_{g_0}$ represent the weight matrix and bias used to produce the corresponding gating signal, and $W_y$ and $b_y$ transform $c$ into logits before feeding to \emph{softmax} activation. 

\begin{figure}[]
	\centering
	\includegraphics[width=0.9\linewidth]{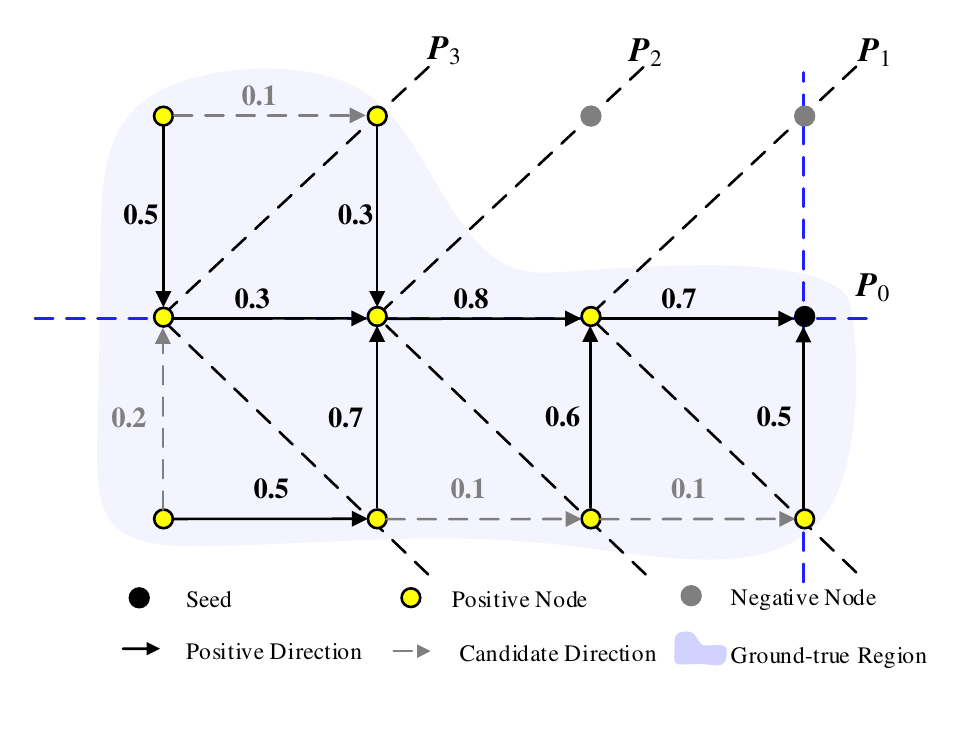}
	\caption{Ground-truth Labeling: All arrows represent the candidate merging directions for nodes, and the corresponding scores predicted by CSE are marked. The candidate directions with highest scores are label as positive directions, which are represented by solid arrows.}
	\label{fig:cse_label}
\end{figure}

\subsection{Seeding} \label{sect:seeding}
In our CSE method, a seed is assumed to be located within an object region. This prerequisite can be easily guaranteed by using the outputs of an off-the-shelf object detector. Here, we adopt detected box centers and shapes by Faster RCNN \cite{ren2015faster} to decide seed locations and the shapes, and uniformly sample $S\times S$ features using bi-linear interpolation from a region indicated by a bounding box. In fact, our CSE method only requires a seed to be located within an object region with sampling grid coarsely covering the targeted object. Moreover, given a sampling grid, any node within the object region can be specified as a seed. As shown in Sect. \ref{sect:flx_and_rob}, randomly initializing seed location and corresponding grid size does not significantly affect the performance. Therefore, a weaker detector, which is easy to optimize, could be sufficient for CSE to produce satisfactory results. \par

\subsection{Optimization}

\textbf{Labeling} In the training phase, the ground-true merging directions are labeled using the strategy illustrated in Fig. \ref{fig:cse_label}. For each grid in CSE, we first label the nodes within the target ground-true object region as positive and the rest as negative. For every positive node, we search its neighborhood positive nodes at the previous section and label the corresponding merging directions as candidate directions\footnote{At most two merging directions will be labeled as positive.}. Among the candidate directions of the same node, we only label the one with the highest score as the final positive merging direction. For the seed node, we always label its $y_s$ as positive.\par 

\textbf{Loss Function} We apply cross-entropy loss to each node to optimize our CSE model, which can be represented by
\begin{align}
& \mathcal{L}_{cse} = \frac{1}{N}\sum_{p \in P} -\ln (y^*(p)),
\end{align}
where $N = S\times S$ represents the number of nodes in a grid, $P$ denotes a set of all nodes, and $y^*$ is the value of the positive merging direction. Our optimization strategy computes the loss according to the current CSE prediction. Intuitively, it intends to boost the positive candidates which are already strong, which reduces the ambiguity in labeling and speed-up the convergence.

\section{Experiment}

\subsection{Experiment Details}
The experiment is conducted on Tensorflow 1.5.0 \cite{abadi2016tensorflow}. We adopt Faster RCNN driven by ResNet-34 to initialize seed locations and corresponding grids in all experiments. The CSE is implemented and optimized in C++ and accelerated by CUDA. Following the existing training strategies for scene text detection \cite{zhou2017east,lyu2018eccv,liu2018fots}, we pretrain our model on a combined dataset. The pretraining dataset consists of over 10k images from full set of ICDAR-17 MLT \cite{icdar2017} and the training sets of MSRA-TD500 \cite{yao2012detecting}, Total-Text \cite{ch2017total} and CTW-1500 \cite{yuliang2017detecting}. After the pretraining, we fine-tune and evaluate our method on two curve text datasets Total-Text (with 1255 training images and 300 testing images) and CTW-1500 (with 1000 training images and 500 testing images). The model is trained on the combined dataset for 50k iteration and fine-tuned on the datasets to be evaluated. We adopt the Adam optimizer \cite{kingma2014adam} to train the network. In the pretraining phase, the learning rate is fixed to $0.01$ for the first 30k iterations and scaled down to $0.002$ for the rest iterations. In the fine-tuning, the initial learning rate is set to $0.001$ and decays exponentially 0.9 every 5000 iterations. All the experiment is conducted on Dell Alienware with Intel i7 processor, 64GB memory and two NVIDIA GTX 1080 Ti GPUs. The batch size is set to 1 for each of two GPUs in training and only one GPU is used for evaluation.

\subsection{Flexibility and Robustness} \label{sect:flx_and_rob}
\begin{figure*}[]
	\centering
	\includegraphics[width=0.9\linewidth]{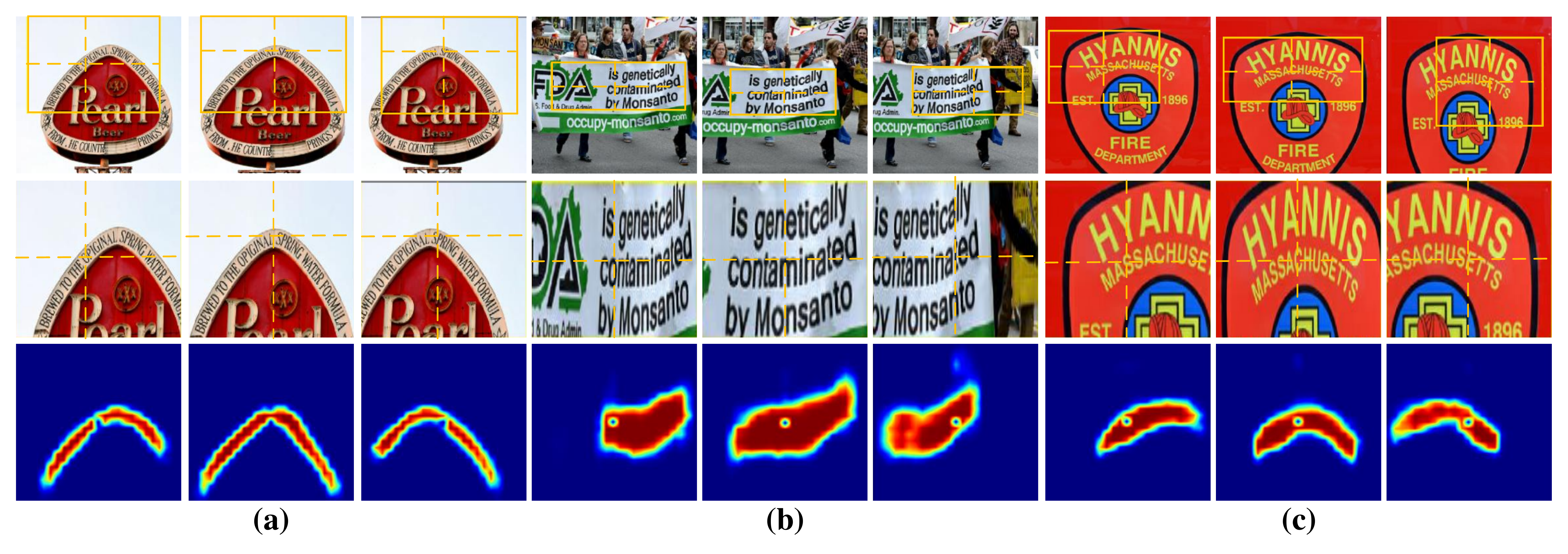}
	\caption{Robustness and Flexibility Analysis: In the first row of each case, the sampling regions are represented by the bounding boxes in yellow and the seed locations are labeled by the cross dash lines. The second row shows the zoom-in of RoIs. The corresponding heatmaps of associate regions are shown in the third row. As shown in (a), our method is very flexible in the seed's locations. (b) and (c) prove the robustness of our method to extract text from a density text region.}
	\label{fig:flex_and_rob}
\end{figure*} 
In this experiment, we validate the flexibility and robustness of our CSE method qualitatively and quantitatively. In the qualitative experiment, we generate a set of sampling grids (in yellow) with different locations and sizes by randomly manipulating the ground-true boxes as shown in the first row of Fig. \ref{fig:flex_and_rob}. We apply the CSE to the corresponding RoIs shown in the second row of Fig. \ref{fig:flex_and_rob}, and visualize the extracted text regions by heat maps in the third row of Fig. \ref{fig:flex_and_rob}. Fig. \ref{fig:flex_and_rob} (a) shows the flexibility of our method. Our CSE method can effectively retrieve the text region with different seed's locations. Even for a text object with large curvature and slim shape, our method can capture all the related sub-regions with high area precision. On the other hand, for a proposed region with many unexpected texts included or even dominated by another text instances (demonstrated in Fig. \ref{fig:flex_and_rob} (b) and (c)), our method only extracts associated object regions indicated by the seed. It indicates that our CSE is robust to the ambiguity caused unexpected objects and can produce satisfactory results even for a poor sampling grid generated by the previous object detector. \par

\begin{figure}[]
	\centering
	\includegraphics[width=0.9\linewidth]{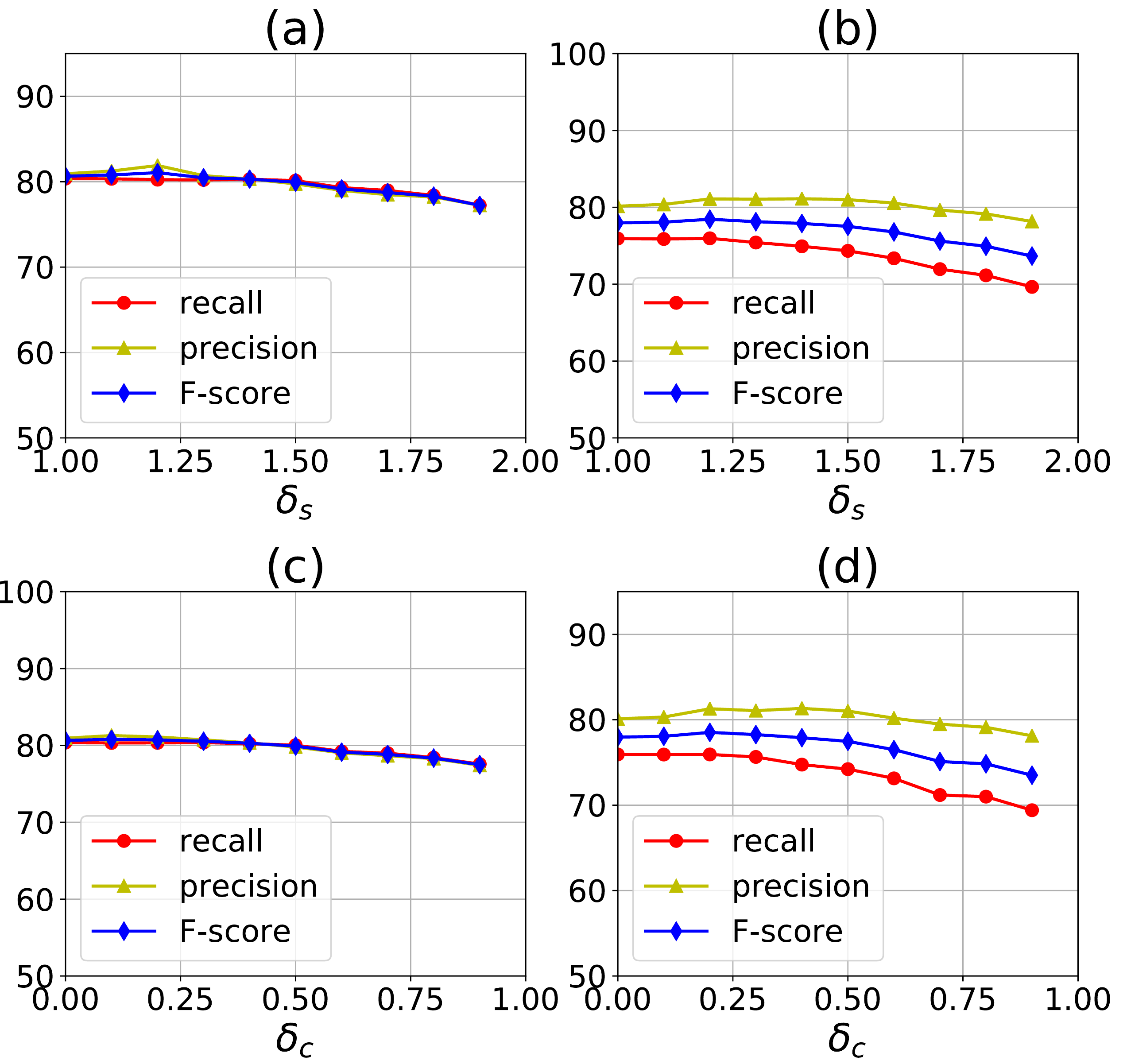}
	\caption{Performance v.s. rescaling factor $\delta_s$ on Total-Text (a) and CTW-1500 (b); Performance v.s. relocation factor $\delta_c$ on Total-Text (c) and CTW-1500 (d).}
	\label{fig:distort}
\end{figure}

In addition to visually investigating our CSE, we quantitatively verify its flexibility and robustness by rescaling the size of a proposed sampling grid and relocating the seed in a gird. The grid rescaling resizes the height and width of a proposed region proposal by a factor of $\delta_s \geq 1.0$. The seed relocation is applied to a sampling grid to change its seed to a new node which is still within the targeted object region but have $\delta_c$ deviation in Euclidean space from the original seed node. $\delta_c$ is normalized by the square root of the original grid area. We study the effects of rescaling and relocation separately by profiling the precisions, recalls, and F-scores on both Total-Text and CTW-1500, and the results are shown in Fig. \ref{fig:distort}. The performance variation respective to rescaling factor $\delta_s$ on two datasets is profiled in Fig. \ref{fig:distort} (a) and (b). The performance is maintained at around 80$\%$ on Total-Text and 78$\%$ on CTW-1500 for $\delta_s$ ranging from 1.0 to 1.5. It slightly drops when $\delta_s$ is larger than 1.5 and the F-scores remain above 77$\%$ and 73$\%$ respectively. As for seed relocation, the according performance variations on Total-Text and CTW-1500 are illustrated in Fig. \ref{fig:distort} (c) and (d). The performance is not greatly affected by the seed relocation. The F-scores remains 80$\%$ and 78$\%$ and decrease by only 3$\%$ and 4$\%$ when $\delta_c$ is changed from 0.0 to 1.0. In conclusion, our CSE is robust to randomly initialized seed locations and distorted sampling grids.

\subsection{Comparing with Mask RCNN}

The baseline Mask RCNN method is implemented based on the method proposed in \cite{lyu2018eccv}. We remove the text recognition branch and only keep the detection and segmentation branches. For a fair comparison, two methods share the same text detector which is based on Faster RCNN architecture. The quantity results are reported in Tab \ref{tab:mrcnn_cse}. Our method is overall better than the baseline method. The F-scores of the baseline method on both datasets are 67.5$\%$ to 67.8$\%$, respectively. In comparison, our CSE performs much better than the Mask RCNN based method by over 10$\%$ with F-score of 80.2$\%$ on Total-Text and 77.4$\%$ on CTW-1500. \par

\begin{figure}[]
	\centering
	\includegraphics[width=\linewidth]{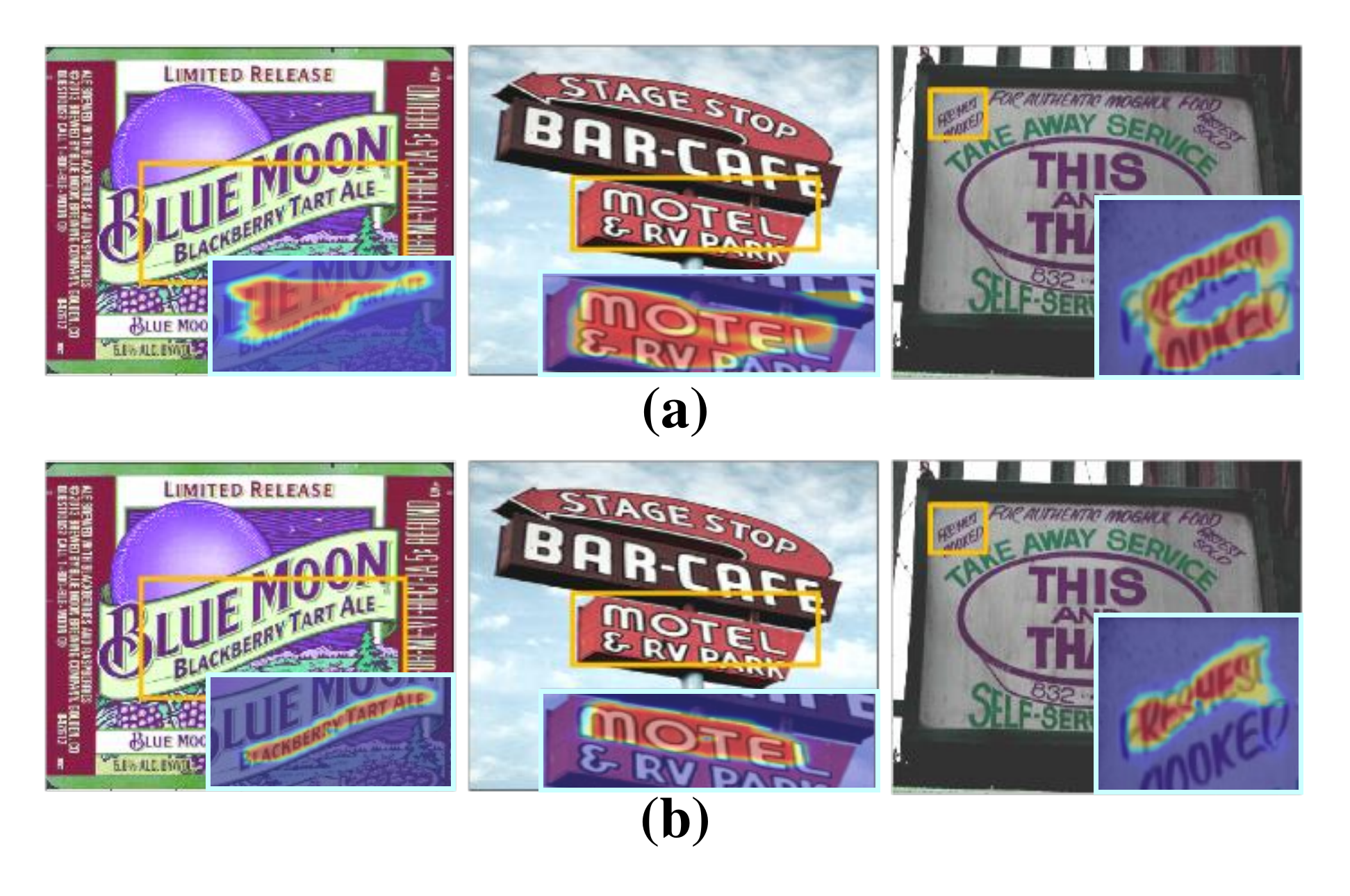}
	\caption{(a) Examples output by baseline method; (b) Examples output by our CSE method. Compared with the baseline Mask RCNN method, our method shows impressing robustness to the ambiguity caused by adjacent texts and impressing selectivity to the targeted instance.}
	\label{fig:demo_mrcnn_cse}
	\vspace{-4mm}
\end{figure}

To explore the cause of performance gain, we visually compare the output confidence map produced by two methods on CTW-1500, which is shown in Fig. \ref{fig:demo_mrcnn_cse}. Fig. \ref{fig:demo_mrcnn_cse} (a) demonstrates the failed examples produced by the baseline method. In these cases, the segmentation is distorted by the adjacent text instance. Parts of the unexpected text instances included in a box cause high activation in a confidence map and corrupt the boundary prediction. In contrast, our CSE is extremely robust in this scenario, since the contextual information captured by CSE helps to eliminate the ambiguity caused by unexpected objects. Moreover, the condition modeling allows our CSE to retrieve long curve text lines with high precision, which is flexible and promising in real applications.
\begin{table}[]\renewcommand{\arraystretch}{1.1}
	\centering
	\label{tab:mrcnn_cse}
	\scalebox{0.95}{
	\begin{tabular}{l||ccc|ccc}
		\hline
		{Datasets}        & \multicolumn{3}{c|}{{Total-Text}} & \multicolumn{3}{c}{{CTW1500}} \\ \hline
		{Model}   & {P}   & {R}  & {F}  & {P}   & {R}  & {F}  \\ \hline
		MRCNN   & 69.2   & 65.8   & 67.5  & 65.1  & 70.8    & 67.8    \\ \hline
		\textbf{CSE}     & \textbf{81.4}   & \textbf{79.1}   & \textbf{80.2}  & \textbf{78.7}  & \textbf{76.1}    & \textbf{77.4}    \\ \hline
	\end{tabular}
	}
	\caption{Performance Comparison between Mask RCNN based method and our CSE method.}
\end{table}

\subsection{Comparing with Polygon Regression}
In this experiment, we compare our method with another baseline based on polygon regression proposed in \cite{yuliang2017detecting}. The baseline is implemented based on the publicly available source code provided in \cite{yuliang2017detecting}, and is pretrained and fine-tuned with our training strategy. Similar to the previous experiment, the backbone and the RPN are shared among two methods, and the rest parts are implemented based on respective workflows. The results are shown in Tab \ref{tab:cmp_poly}. Our method outperforms the baseline method by $5\%$ in terms of F-score on Total-Text, with the precision of 80.9$\%$, recall of 80.3$\%$ and F-score of 80.6$\%$. On CTW-1500, our CSE achieves F-score of 77.6 $\%$, which is 4.4$\%$ better than the baseline.  

\begin{figure}[]
	\centering
	\includegraphics[width=\linewidth]{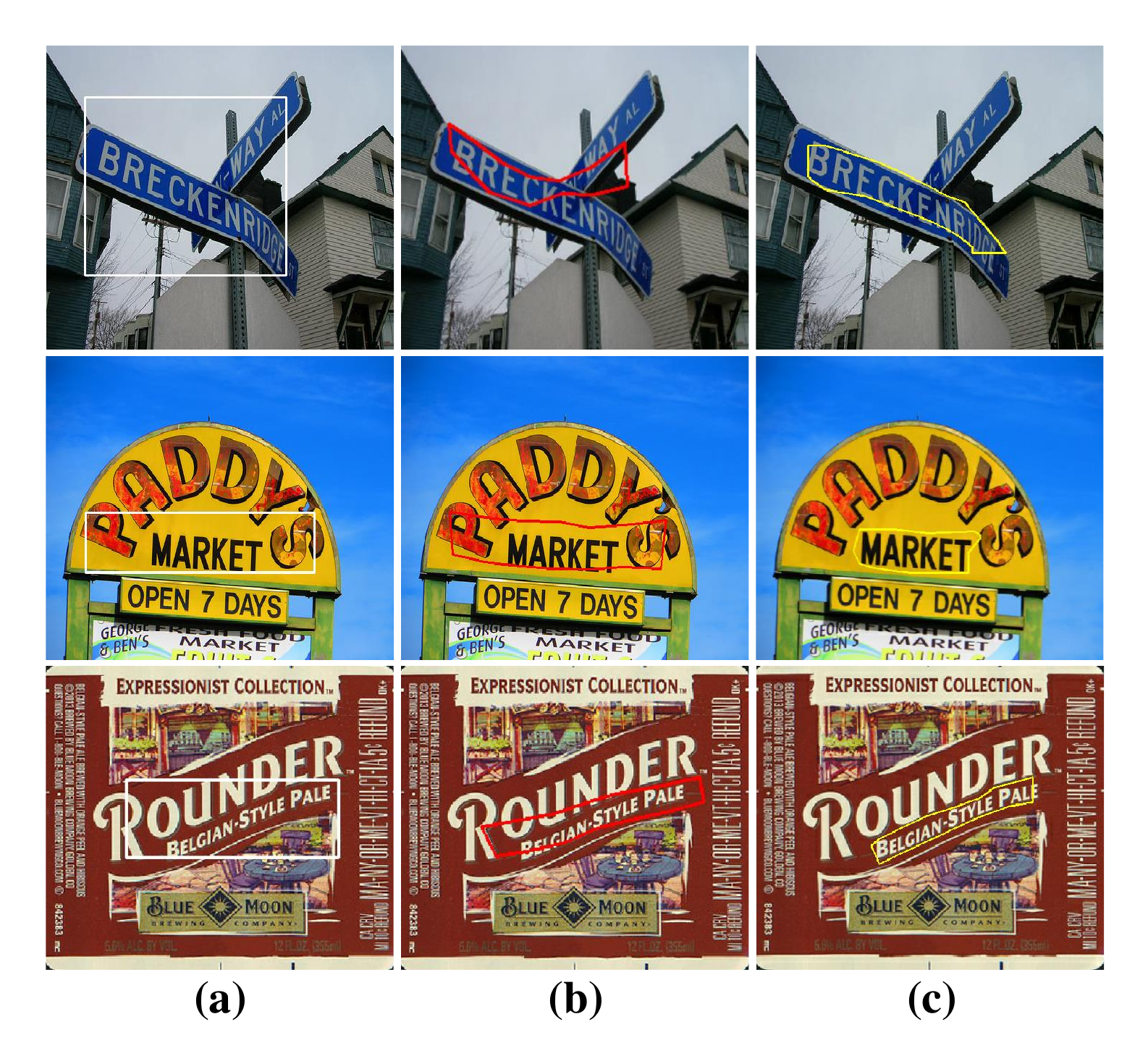}
	\caption{(a) The images with RoIs; (b) The text boundaries output by baseline method; (c) The text boundaries produced by our CSE method. The baseline method is easily affected by the unexpected texts included in the same boxes, while our method shows strong robustness to this situation.}
	\label{fig:demo_poly_cse}
\end{figure}

We also investigate the causes of performance gain by visualizing the detection results of both methods. As demonstrated in Fig. \ref{fig:demo_poly_cse}, with the same RoI proposals shown in Fig. \ref{fig:demo_poly_cse} (a), the polygon regression could be corrupted by the other text object which is occasionally included. As can be seen in Fig. \ref{fig:demo_poly_cse} (b), the baseline model may consider all the texts included in a proposed region as a single object and regress the corresponding boundary. Although this problem can be mitigated by training a more accurate text detector to reduce the unexpected texts, the proposed bounding boxes inevitably cover additional texts due to text's highly varying shapes and orientations. As shown in Fig. \ref{fig:demo_poly_cse} (c), the conditional expansion mechanism only merges the sub-regions that are similar to the region indicated by the seed. By exploring the spatial dependency as well as the local information, our CSE method is much more robust than the polygon regression method and can produce more elaborated boundaries. \par
\begin{table}[]\renewcommand{\arraystretch}{1.1}
	\centering
	\label{tab:cmp_poly}
	\scalebox{0.9}{
	\begin{tabular}{l||ccc|ccc}
		\hline
		{Datasets}        & \multicolumn{3}{c|}{{Total-Text}} & \multicolumn{3}{c}{{CTW1500}} \\ \hline
		{Methods}   & {P}   & {R}  & {F}  & {P}   & {R}  & {F}  \\ \hline
		Poly-Reg  & 73.8   & 77.4   & 75.6    & 77.1   & 69.7   & 73.2       \\ \hline
		\textbf{CSE}  	  & \textbf{80.9}   & \textbf{80.3}   & \textbf{80.6}    & \textbf{79.2}   & \textbf{76.0}   & \textbf{77.6}    \\ \hline
	\end{tabular}
	}
\caption{Performance Comparison between Polygon Regression based method and our CSE method.}
\end{table}

\subsection{Peer Comparison}
\begin{figure}[]
	\centering
	\includegraphics[width=0.9\linewidth]{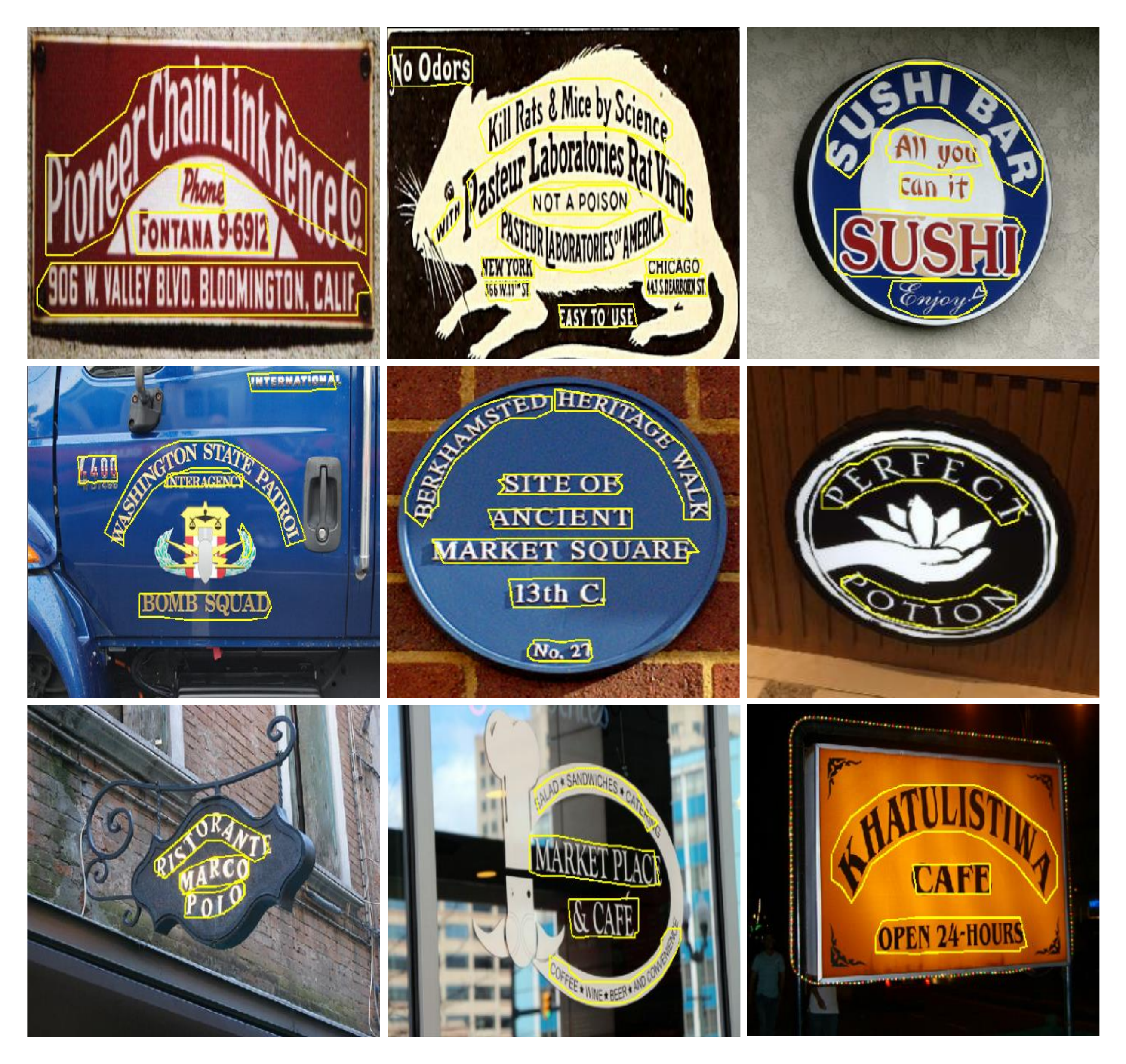}
	\caption{Detection results on Total-Text and CTW-1500.}
	\label{fig:demo_all}
\end{figure}

We compare our method with the recently proposed methods for curve text detection on curve text benchmarks, Total-Text and CTW-1500. The results are shown in Tab.\ref{tab:cmp_tt} and Tab.\ref{tab:cmp_ctw}. Our method creates a new state-of-art performance with the precision of 81.4$\%$, recall of 79.1$\%$ and F-score of 80.2$\%$ on Total-Text. On CTW-1500 containing both curve texts and long text lines, our method also achieves the state-of-art performance with the precision of 81.1$\%$, recall of 76.0$\%$ and F-score of 78.4$\%$. The inference time is 0.42 ms per image and 0.38 ms per image on Total-Text and CTW-1500 respectively. The detection results are demonstrated in Fig. \ref{fig:demo_all}. It shows that our method can effectively handle curve texts with irregular shapes, highly varying sizes and arbitrary orientations.
\begin{table}[]\renewcommand{\arraystretch}{1.2}
\centering
\label{tab:cmp_tt}
\scalebox{0.9}{
\begin{tabular}{l||cccc}
	\hline
	Methods       & P & R & F & time (s)        \\ \hline
	SegLink  \cite{shi2017detecting} & 30.3  & 23.8 & 26.7 & -\\
	EAST \cite{zhou2017east}   & 50.0 & 36.2 & 42.0 & - \\                 
	Mask TextSpotter \cite{lyu2018eccv} & 69.0 & 55.0 & 61.3 & - \\ 
	TextSnake et al. \cite{long2018eccv} & \textbf{82.7} & 74.5 & 78.4 & - \\ \hline
	\textbf{CSE} & 81.4 & \textbf{79.1} & \textbf{80.2}  & 0.42\\ \hline
\end{tabular}
}
\caption{Detection Performance on Total-Text.}
\end{table}
\begin{table}[]\renewcommand{\arraystretch}{1.2}
\centering
\label{tab:cmp_ctw}
\scalebox{0.9}{
\begin{tabular}{l||cccc}
	\hline
	Methods       & P & R & F & time (s)        \\ \hline
	SegLink  \cite{shi2017detecting}      & 42.3 & 40.0 & 40.8 & - \\
	EAST \cite{zhou2017east}   & 78.7 & 49.1 & 60.4 & - \\      
	DMPNet \cite{liu2017deep} & 69.9 & 56.0 & 62.2 & - \\
	CTD \cite{yuliang2017detecting} & 74.3 & 65.2 & 69.5 & - \\
	CTD+TLOC \cite{yuliang2017detecting} & 77.4 & 69.8 & 73.4 & - \\ 
	TextSnake et al. \cite{long2018eccv} & 67.9 & 85.3 & 75.6 & - \\ \hline
	\textbf{CSE} & \textbf{81.1} & \textbf{76.0} & \textbf{78.4}  & 0.38\\ \hline
\end{tabular}
}
\caption{Detection Performance on CTW-1500.}
\vspace{-4mm}
\end{table}

\section{Conclusion}

In this work, we analyze the deficiency of the existing curve text detection methods and improve the performance by developing a novel parameterized Conditional Spatial Expansion (CSE) mechanism. Our method shows strong robustness to the ambiguity caused by close texts with arbitrary shapes and orientations. It is flexible and can extract text regions in a controllable manner. Our CSE method outperforms the existing curve text detection methods.
\section*{Acknowledgement} 
G. Lin's participation was partly supported by the National Research Foundation Singapore under its AI Singapore Programme [AISG-RP-2018-003] and a MOE Tier-1 research grant [RG126/17 (S)].

{\small
\bibliographystyle{ieee_fullname}
\bibliography{egbib}
}

\balance
\appendix

\section{Implementation Details}

\subsection{System Architecture}

\subsubsection{Backbone}
\begin{figure}[th]
	\includegraphics[width=0.5\textwidth]{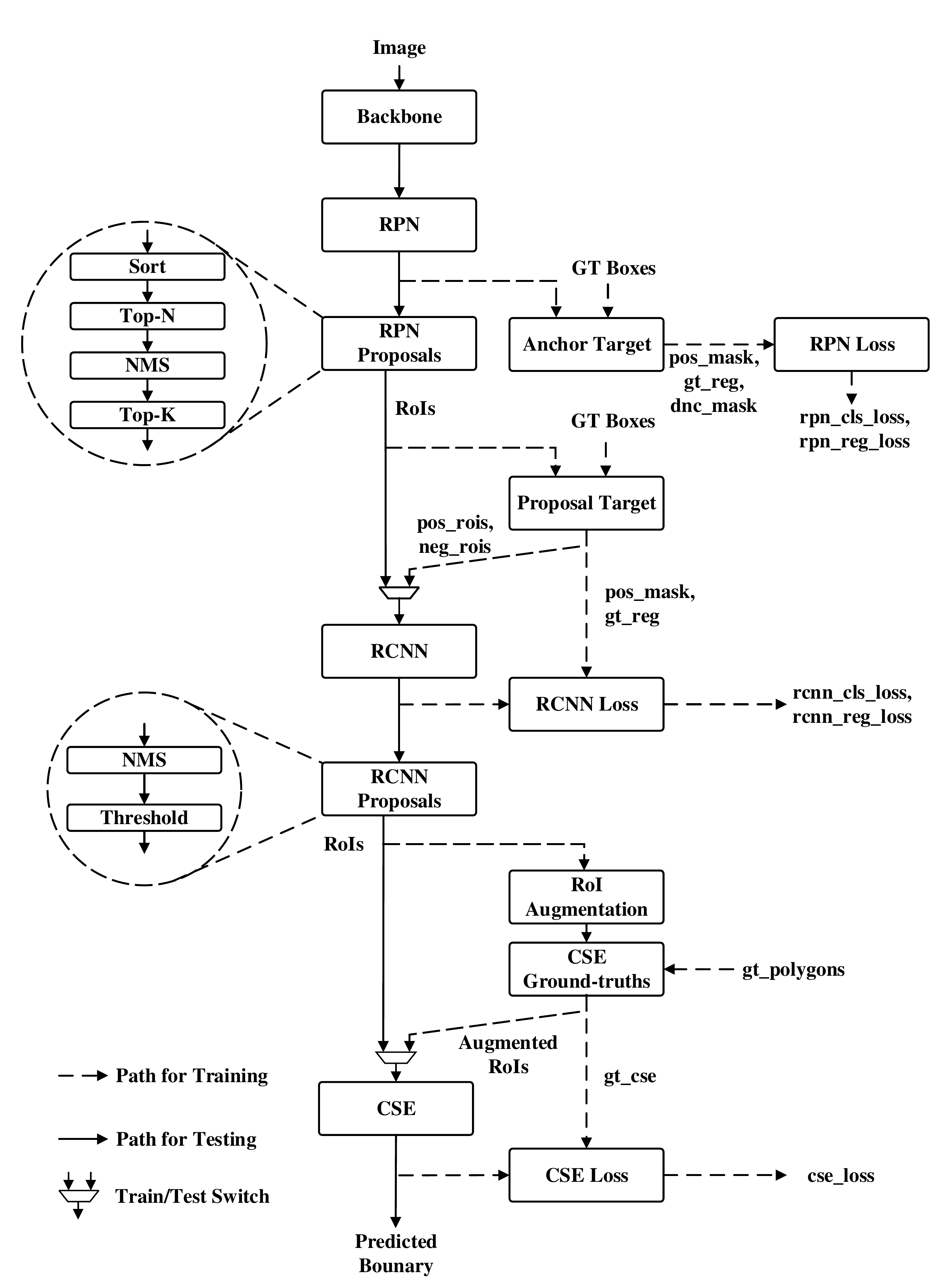}
	\centering
	\caption{System Architecture.}
	\label{fig:arch}
\end{figure}

The system architecture is illustrated in Fig. \ref{fig:arch}. It consists of a backbone network, a faster RCNN and our CSE module. The backbone network is used to encode an image into spatial features, which is composed of a ResNet-34 \cite{resnet34} and a Feature Pyramid Network (FPN) \cite{lin2017feature}. For the ResNet-34, we remove the fully-connected layers and keep the rest with output feature dimension of 512 and a down-sampling factor of $\mathcal{Q}=1/16$. Following the ResNet-34, we apply an FPN with additional 5 down-sampling layers and $N_u$ up-sampling layers. The number of output channel of each down-sampling or up-sampling layer is 512. Notably, each up-sampling layer up-samples an input feature map with a factor of 2, and $N_u$ varies with the targeted benchmarks where the evaluation is conducted. For the curve text benchmarks \cite{ch2017total,yuliang2017detecting}, $N_u$ is set to 5 to produce an output feature map with down-sampling factor of $1/16$. For the non-curve text benchmarks \cite{karatzas2015icdar}, the optimal down-sampling is heuristically set to $1/4$ following the existing approaches \cite{lyu2018eccv,long2018eccv,yuliang2017detecting,ma2018arbitrary,zhou2017east}, and thus two additional up-sampling layers are applied. \par

\subsubsection{Faster RCNN for Seeding}
The Faster RCNN \cite{ren2015faster} is built on top of the backbone to initialized the seeds' locations and corresponding grid sizes. The features computed by backbone network are first fed to the Region Proposal Network \cite{ren2015faster} to generate coarse region proposals. In the RPN, 20 anchors with 5 scales (32, 64, 128, 256, 512) and 4 aspect ratios (0.25, 0.5, 1, 2) are defined. In RoI generation, we first select 6k anchors with highest scores before NMS and output 300 RoIs after that. Here, the IoU threshold is set to be 0.7. The proposed RoIs are subsequently input to the RCNN \cite{girshick2015fast} for fore/background classification and bounding box calibration. In this stage, we suppress the overlapped bounding boxes using NMS with IoU threshold of 0.5. The reason for setting a high IoU threshold is that the bounding boxes are highly overlapping if their corresponding text object is close to each other. Setting a low IoU threshold, e.g. 0.3, will incorrectly remove these objects. After the second NMS, we remove the box proposals with positive scores lower than 0.7 and feed the resulted box proposals to CSE module. \par 

\subsubsection{CSE}
Our CSE takes the output boxes by Faster RCNN to generate a set of $S\times S$ sampling grids. These grids represent the sampling locations on the original images, which will be used to compute the values of sampled features from the output of the backbone based on bilinear interpolation method. Here, $S$ is set to 25. Subsequently, the CSE is applied to the sampled feature points to extract the text region indicated by a seed. Finally, the instance-level boundaries are produced by mapping associated feature points back to the input image and then extracted corresponding contours. \par

\subsection{Training}

\subsubsection{Faster RCNN}
The training of Faster RCNN follows the standard training strategy proposed in \cite{ren2015faster,shrivastava2016training}. The short sides of the input images are fixed to 512 pixels and the aspect ratios are configured to be lower than 1.5. Color distortion and rotation are applied to augment the training images. Since the size of texts' ground-true boxes is highly varying, the original threshold 0.7 easily leads to a batch overwhelmed by negative samples, which hinders the convergence. We maintain the batch size to be 256 but reduce the positive matching threshold of RPN to 0.6 to maintain a sufficient number of positive anchors.  As for the training of RCNN, the batch size is set to 128 and the positive RoI threshold is set to 0.5. Additionally, we apply the Online Hard Example Mining (OHEM) \cite{shrivastava2016training} to balance the number of positives and negatives to 1:3. 

\subsubsection{CSE}
A batch to train CSE is configured to have 10 samples with ground-true seeds and grids equally sampled from the positive RoI produced by previous Faster RCNN and the augmentations of the groud-true text boxes. The augmented seeds and grids are constructed by manipulating the grid sizes and the seeds' locations under the constraints that an augmented grid should have at least 0.4 overlapping with the ground-true boxes and a seed should be within the targeted object region. After the RoI augmentation, we compute a ground-true foreground mask for each RoI based on the ground-true polygons provided in the datasets. The ground-true foreground mask will be used to compute the CSE loss for optimization.

\section{Performance on Non-curve Text Benchmarks}
We evaluate the performance of our method to detect non-curve texts on ICDAR-2015 \cite{karatzas2015icdar}. As illustrated in Tab. \ref{tab:cmp_ic15}, our method shows impressive performance on non-curve text dataset containing texts with various shapes and orientations. Our CSE method achieves the state-of-art performance with a precision of 92.3, recall of 79.9 and F-score of 85.7.\par

\begin{table}[h]\renewcommand{\arraystretch}{1.2}
\centering
\caption{Localization performance on ICDAR 2015.}
\label{tab:cmp_ic15}
\scalebox{1}{
\begin{tabular}{l||ccc}
	\hline
	Methods       & P & R & F         \\ \hline
	DeepReg \cite{he2017deep}         & 82.0 & 80.0 & 81.0 \\
	EAST \cite{zhou2017east} & 83.3 & 78.3 & 80.7 \\
	R$^2$CNN \cite{jiang2017r2cnn}    & 85.0 & 80.0 & 82.4 \\
	RRPN  \cite{ma2018arbitrary}      & 84.0 & 77.0 & 80.3 \\  
	Mask TextSpotter. \cite{lyu2018eccv} & 88.7 & 80.1 & 84.1 \\   \hline
	CSE  & \textbf{92.3} & 79.9  & \textbf{85.7} \\  \hline

\end{tabular}
}
\end{table}

\end{document}